\documentclass[final,5p,times,twocolumn]{elsarticle}

\usepackage{amssymb}
\usepackage{amsmath}
\usepackage{algorithm}
\usepackage{algorithmic}
\usepackage{stfloats}
\usepackage{subcaption}
\usepackage{etoolbox}
\usepackage{multirow}
\usepackage{graphicx} % include this in your preamble

\usepackage{booktabs} % For formal tables
\usepackage{tabularx} % For full-width tables
\usepackage{siunitx} % For alignment of numbers
\usepackage{makecell} % For line breaks in cells
\usepackage{caption} % For customizing captions
\usepackage{url}
\newcolumntype{Y}{>{\centering\arraybackslash}X}

\makeatletter
\newcommand{\INPUT}[1]{\ALC@it\algorithmicinput\ #1 }
\newcommand{\algorithmicinput}{\textbf{Input:}}
\makeatother

\begin{document}

\begin{frontmatter}



\title{Cross-Modal Augmentation for Few-Shot Multimodal Fake News Detection}


%\corref{cor1}
\author[label1]{Ye Jiang}
\author[label1]{Taihang Wang}
\author[label1]{Xiaoman Xu}
\author[label2]{Yimin Wang\corref{cor1}}
\author[label3]{Xingyi Song}
\author[label3]{Diana Maynard}

\affiliation[label1]{organization={College of Information Science and Technology},%Department and Organization
            addressline={ Qingdao University of Science and Technology}, 
            country={China}}

\affiliation[label2]{organization={College of Data Science},%Department and Organization
            addressline={ Qingdao University of Science and Technology}, 
            country={China}}     
\cortext[cor1]{Corresponding author}

\affiliation[label3]{organization={Department of Computer Science},%Department and Organization
            addressline={ University of Sheffield}, 
            country={UK}}
            
\begin{abstract}
The nascent topic of fake news requires automatic detection methods to quickly learn from limited annotated samples. Therefore, the capacity to rapidly acquire proficiency in a new task with limited guidance, also known as few-shot learning, is critical for detecting fake news in its early stages. Existing approaches either involve fine-tuning pre-trained language models which come with a large number of parameters, or training a complex neural network from scratch with large-scale annotated datasets. This paper presents a multimodal fake news detection model which augments multimodal features using unimodal features. For this purpose, we introduce Cross-Modal Augmentation (CMA), a simple approach for enhancing few-shot multimodal fake news detection by transforming n-shot classification into a more robust (n $\times$ z)-shot problem, where z represents the number of supplementary features. The proposed CMA achieves SOTA results over three benchmark datasets, utilizing a surprisingly simple linear probing method to classify multimodal fake news with only a few training samples. Furthermore, our method is significantly more lightweight than prior approaches, particularly in terms of the number of trainable parameters and epoch times. The code is available here: \url{https://github.com/zgjiangtoby/FND_fewshot}

\end{abstract}


% \begin{highlights}
% \end{highlights}


\end{frontmatter}

\section{Introduction}
The recent proliferation of social media has not only transformed the landscape of information exchange, but also led to the pernicious spread of fake news. The detection and mitigation of fake news have consequently become pivotal areas of research \cite{conroy2015automatic, long-etal-2017-fake}. Traditional approaches, primarily relying on textual analysis, have shown limitations due to the sophisticated and multi-faceted nature of fake news \cite{wang2018eann, lao2021rumor}. In response, many studies have incorporated multimodal methods that consider both text and accompanying images, yielding a more comprehensive and effective framework for identifying and debunking fake news \cite{chen2022cross, zhou2023multimodal}.

To explore the inconsistent semantics between text and image in fake news, many studies have either incorporated contrastive learning to achieve better alignment between image-text pairs \cite{wang2023cross}, or designed complex neural networks to strengthen the deep-level fusion of multimodal features \cite{wu2023mfir, qu2024qmfnd}. The former relies on contrastive loss to align image-text pairs, but most image-text pairs in fake news are inherently not matched \cite{gao2022pyramidclip}, and different image-text pairs may also have potential correlations \cite{li2021align}, which can consequently confuse the model. The latter typically needs to be trained from scratch, which is fundamentally bounded by the availability of large-scale annotated data \cite{rashkin2017truth, shu2020fakenewsnet}. 

\begin{figure}[h!]
	\centering
		\includegraphics[scale=.5]{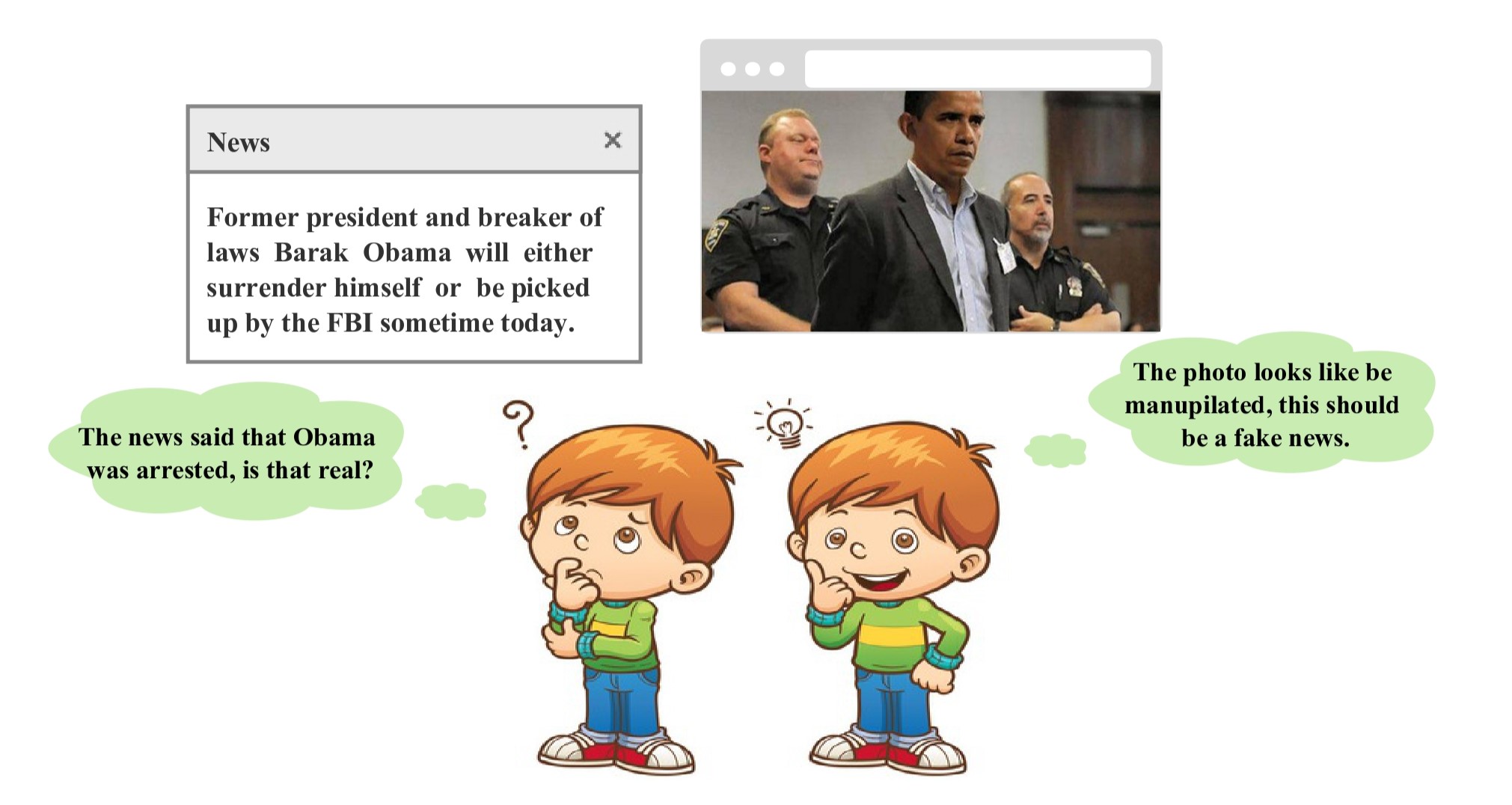}
	\caption{Information from different modalities assists humans in decision-making, especially when faced with uncertainty.}
	\label{FIG:fake}
\end{figure}

In contrast to machines, the process of concept learning in humans involves integrating multimodal signals and representations \cite{meltzoff1979intermodal, nanay2018multimodal}. When processing uncertain information, people inherently seek help from other modalities. This capability enables humans to learn from a limited number of samples by incorporating cross-modal information, as shown in Figure \ref{FIG:fake}. Meanwhile, the efficacy of fake news detection (FND) in the context of nascent topics, such as COVID-19, remains a significant challenge for prevailing strategies. This difficulty is compounded by the lack of extensive data and annotations in the target domain, underscoring the critical role of few-shot learning in mitigating the spread of early-stage fake news \cite{wu2023prompt}. 

% Existing approaches such as fine-tuning\cite{}, prompting\cite{}, or encoding\cite{} from the pre-trained language models (PLMs), are increasingly being utilized in the task of fake news detection (FND). However, the efficacy of FND in the context of nascent topics (e.g., COVID-19), remains a significant challenge for prevailing strategies. This difficulty is compounded by the lack of extensive data and annotations from the target domain, underscoring the critical role of few-shot learning in mitigating early-stage fake news detection\cite{}.

% which impedes the ability to accurately identify fake news, thereby

In the context of emerging topics with limited training samples, prompt learning, through its few-shot learning capacity, encapsulates news articles in task-specific textual prompts for direct knowledge extraction from pre-trained language models (PLMs), achieving comparable performance across different tasks \cite{gao2021making, ding2021openprompt}. However, most prompt-based methods primarily tune the PLM with unimodal textual information from fake news \cite{jiang2022fake, wu2023prompt}, thus once again ignoring the multimodal nature of fake news. Even though the previous method \cite{JIANG2023119446} attempts to integrate the different prompt templates with image features extracted from the pre-trained vision model, the fusion strategy still utilized the multimodal features only, potentially struggling to address spatial discrepancies between visual and textual semantics \cite{wu2023mfir, guo2023tiefake}.

% When processing false information visually, the cerebral cortex activates neurons similarly as it would when encountering fake news through other modalities such as linguistic channels \cite{gibson1969principles, meltzoff1979intermodal, nanay2018multimodal}. argue that the FND model should be capable of quickly learning a new task with minimal instructions by using multimodal data, and

In this paper, we propose a \textbf{C}ross-\textbf{M}odal \textbf{A}ugmentation (CMA) method to explore how unimodal features could assist in multimodal fusion for FND in few-shot scenarios. Specifically, we leverage the foundational multimodal model CLIP \cite{radford2021learning} to extract textual and visual features from fake news simultaneously. Utilizing class labels as supplementary one-shot training instances, the n-shot classification can then be converted to an $(n \times z)$-shot problem, where $z$ represents the number of supplementary features (e.g., the fused feature from text and image). Meanwhile, we also fuse unimodal features by utilizing the cross-attention mechanism \cite{wang2023cross} as another supplementary. Finally, we employ a simple linear probing for each modality as well as for the fused multimodal features. The experimental results indicate that CMA achieves SOTA results across three datasets.

% by taking advantage of additional training samples from each modality

The main contributions of this paper are:

\begin{itemize}
    \item Introduction of a Cross-Modal Augmentation (CMA) method for few-shot multimodal fake news detection, utilizing unimodal features to enhance multimodal fusion.
    
    \item Leveraging a pre-trained multimodal model to extract unimodal features, and repurposing class labels as additional one-shot training samples, transforming the n-shot classification into a more robust \((n \times z)\)-shot problem.
    
    \item By freezing the pre-trained multimodal model and training only with a simple linear classifier, the proposed CMA achieves SOTA results over three datasets, outperforming 11 baseline models and surpassing previous methods in efficiency.
\end{itemize}

\section{Related work}

\subsection{Unimodal fake news detection}

Unimodal fake news detection aims to extract significant semantics from either news texts or images. Given the precision of semantics in text, previous approaches have concentrated on the task of text-based unimodal fake news detection. Early works focused on analyzing statistical characteristics of text (e.g., length, punctuation, exclamation marks) \cite{castillo2011information} and metadata (e.g., likes, shares) \cite{tabibian2017distilling, geeng2020fake} for manual fake news detection. However, these manual feature engineering approaches are time-consuming and struggle with processing large-scale, real-time data \cite{liu2016reuters, fedoryszak2019real}.

The advent of deep learning has significantly advanced automated fake news detection. These methods primarily utilize deep learning models like BiLSTM \cite{bahad2019fake, sridhar2021fake}, GNNs \cite{phan2023fake}, and pre-trained models (e.g., BERT, GPT) \cite{song2021classification, jiang2021categorising, jiang2020comparing} to analyze text features, extracting various attributes such as emotional \cite{ghanem2020emotional}, stance-based \cite{jiang-2023-team}, and stylistic elements \cite{wu2021multimodal}. However, the recent proliferation of multimodal information (text, images, videos) in social networks has shifted the propagation of fake news from solely text-based to multimodal formats.

\begin{figure*}[h!]
	\centering
		\includegraphics[width=\textwidth]{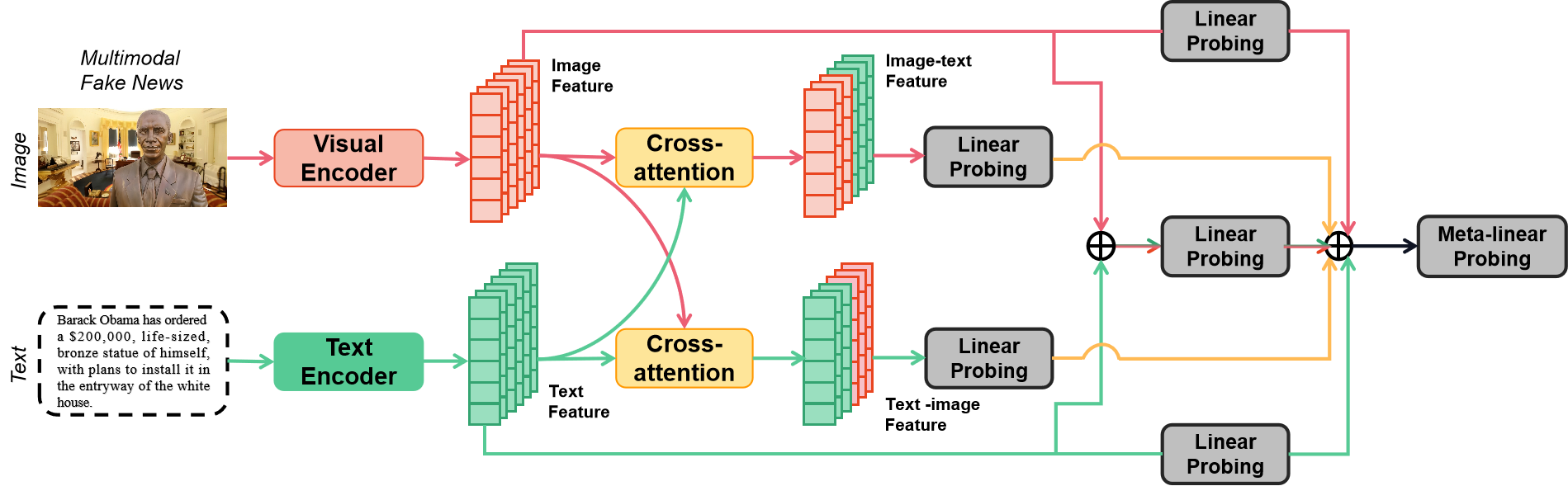}
	\caption{The overall architecture of the CMA model.}
	\label{FIG:model}
\end{figure*}

\subsection{Multimodal fake news detection}

% Unimodal fake news detection aims to extract significant semantics from either news texts or images.  Early works focused on analyzing statistical characteristics of text (e.g., length, punctuation, exclamation marks) \cite{castillo2011information} and metadata (e.g., likes, shares) \cite{tabibian2017distilling, geeng2020fake} for manual fake news detection. Recently, FND methods primarily utilize deep learning models like BiLSTM \cite{bahad2019fake, sridhar2021fake}, GNNs \cite{phan2023fake}, and pre-trained models (e.g., BERT, GPT) \cite{song2021classification, jiang2021categorising, jiang2020comparing} to analyze text features, extracting various attributes such as emotional \cite{ghanem2020emotional}, stance-based \cite{jiang-2023-team}, and stylistic elements \cite{wu2021multimodal}. However, the recent proliferation of multimodal information (text, images, videos) in social networks has shifted the propagation of fake news from solely text-based to multimodal formats.

% To address the above issue, various methods

Multimodal methods employing cross-modal discriminative patterns have been introduced, aiming to enhance performance in fake news detection. For example, MCAN \cite{wu2021multimodal} employs multiple co-attention layers to more effectively integrate textual and visual features in detecting fake news. CAFE \cite{chen2022cross} quantifies cross-modal ambiguity through the assessment of the Kullback-Leibler (KL) divergence among the distributions of unimodal features. LIIMR \cite{singhal2022leveraging} determines the modality that exhibits greater confidence in the context of fake news detection. COOLANT \cite{wang2023cross} focuses on improving the alignment between image and text representations, utilizing contrastive learning for finer semantic alignment and cross-modal fusion to learn inter-modality correlations. However, these approaches are limited by the need for extensive annotated data in the context of emerging topics.

% EANN \cite{wang2018eann} utilizes an auxiliary event discriminator to facilitate the extraction of cross-modal features. MVAE \cite{khattar2019mvae} implements a multimodal variational autoencoder for learning probabilistic latent variable models, subsequently reconstructing original texts and low-level image attributes.

\subsection{Cross-modal few-shot fake news detection}

Few-shot learning is designed to master new tasks using a limited number of labeled examples \cite{wang2020generalizing}. Current few-shot learning methodologies, such as prototypical networks, acquire class-specific features in metric spaces for swift adaptation to novel tasks \cite{vinyals2016matching, snell2017prototypical}. Within computer vision, the concept of few-shot domain adaptation is explored in image classification for transferring knowledge to novel target domains \cite{motiian2017few, zhao2021domain}. In natural language processing, meta-learning is suggested as a means to enhance few-shot learning performance in tasks like language modeling \cite{sharaf2020meta, han2021meta} and misinformation detection \cite{zhu2023metaadapt, zhang2021learning}. To our knowledge, the application of few-shot multimodal fake news detection through cross-modal augmentation remains unexplored in existing literature.

Meanwhile, previous multimodal learning approaches have sought to enhance unimodal tasks by leveraging data from various modalities \cite{schwartz2022baby, zhang2021cross}. With multimodal pre-trained models achieving notable success in classic vision tasks \cite{radford2021learning, zhang2022glipv2}, there is a growing interest in formulating more efficient cross-modal augmentation techniques.

However, the prevailing techniques are based on successful strategies originally designed for multimodal foundational models. For example, CLIP utilizes linear probing \cite{he2022masked, he2020momentum} and comprehensive fine-tuning \cite{girdhar2017attentional} in its application to downstream tasks. CLIP-Adapter \cite{gao2023clip} and Tip-Adapter \cite{zhang2021tip} draw inspiration from parameter-efficient finetuning approaches \cite{houlsby2019parameter} that focus on optimizing lightweight MLPs while maintaining a fixed encoder. However, all the aforementioned methods, including WiSE-FT \cite{wortsman2022robust}, employ an alternative modality, such as textual labels, as classifier weights, and continue to compute a unimodal Softmax loss on few-shot tasks. In contrast, this paper demonstrates the enhanced effectiveness of incorporating additional modalities as training samples.

\section{Methodology}

The proposed CMA enhances few-shot fake news detection by integrating samples from different modalities, and extends traditional unimodal few-shot classification to leverage the richness of cross-modal data, as shown in Figure \ref{FIG:model}.

This section starts with a standard unimodal few-shot FND framework, and the loss function is discussed. Then, it extends this to multiple modalities, assuming each training example is a combination of five different modalities. The modality-specific features are passed through MLP linear classifiers to obtain their inferences. Finally, we combine the inferences and train a meta-linear classifier to compute the final prediction.

\subsection{Unimodal few-shot FND}

Initially, unimodal few-shot FND learns from a labeled dataset of $(x, y) \in X$, where $x$ is either the text or image passing to a pre-trained feature encoder $\phi(\cdot)$. The ultimate goal is to allocate a binary classification label of $y \in \{0,1\}$, in which 0 denotes real news and 1 denotes fake news. We assume only an n-shot subset $(x_i, y_i)$ from $X$ is provided for training, where $i \in [1, n]$ (i.e., $n$ samples per class); the rest of $X$ is used as the test set.

Therefore, the standard unimodal FND can be denoted as minimizing the cross-entropy loss $L$:

\begin{equation}
\label{eqa:1}
    L = -(y_{i}log(y_i') + (1-y_{i})log(1-y'_i))
\end{equation}
where $y_i'$ is the model inference from the linear classifier $MLP$ after softmax.

\begin{equation}
\label{eqa:2}
    y_i' = softmax(MLP(f(x_i)) = -log(\frac{e^{w_y * f}}{\sum_{y'} e^{w_{y'} * f}})
\end{equation}
where $f$ is the feature representation from an MLP layer after the unimodal feature encoder, and $w_y$ and $w_{y'}$ are the weights of the ground truth label and the predicted label, respectively.

\subsection{Multimodal few-shot FND}
To extend to multimodal FND, we assume that for each training sample, $f$ is a combination of five feature representations: 1) a text-only feature $f_t$; 2) an image-only feature $f_m$; 3) concatenation of L2 normalized $f_c = [f_t \oplus f_m]$, where $\oplus$ is the concatenation operation; 4) an image-text cross-attended feature $f_{mt}$; 5) a text-image cross-attended feature $f_{tm}$. The cross-attention mechanism, which swaps the text query $Q_t$ with the image query $Q_m$, to obtain the cross-attended feature $f_{mt}$ is denoted as follows:

\begin{equation}
\label{eqa:3}
    f_{mt} = CrossAtt_{m \rightarrow t}(Q_m, K_t, V_t) = softmax(\frac{Q_m K_{t}^{T}}{\sqrt{d}} )V_t
\end{equation}

In contrast, by swapping the image query $Q_m$ with the text query $Q_t$, the cross-attended feature $f_{tm}$ can be obtained:

\begin{equation}
\label{eqa:4}
     f_{tm} = CrossAtt_{t \rightarrow m}(Q_t, K_m, V_m) = softmax(\frac{Q_t K_{m}^{T}}{\sqrt{d}} ) V_m
\end{equation}
where $K_t$ and $K_m$ represent the key vectors for text and image features respectively, $V_t$ and $V_m$ denote the corresponding value vectors, and $d$ refers to the dimensionality of the model.

For the sake of simplification, we assume that the number of $z$ different types of features are considered as distinct modalities. Therefore, each modality can be processed through the linear classifier MLP in the unimodal learning approach, as discussed above, to obtain five inferred probabilities.

\begin{algorithm}

\caption{Cross-modal Augmentation Algorithm}
\begin{algorithmic}[1] % The number tells where the line numbering should start
\INPUT{source data $X$, number of seeds $S$, number of shots $n$}
\STATE Initialize pre-trained multimodal model;
\FOR{seed $\in \{1,2, \dots, S\}$}
    % \STATE Sample $x_n$ and $y_n$ from $X$;
    \FOR{ $x_i$ in $ \{x_1,\dots, x_n\}$}
        \STATE Extract image feature $f_m$ from the pre-trained vision model;
        \STATE Extract text feature $f_t$ from the pre-trained language model;        
        \STATE Concatenate $f_m$ with $f_t$ and L2 normalize to obtain $f_c$;
        \STATE Compute cross-attended features $f_{mt}$ and $f_{tm}$ with Equations \ref{eqa:3} and \ref{eqa:4};
        \STATE Obtain inferences of each of the above features with linear classifiers from Equation \ref{eqa:2};
        \STATE Concatenate inferences and compute the final prediction with Equation \ref{eqa:5};
        \STATE Compute cross-entropy loss with Equation \ref{eqa:1};
    \ENDFOR
\ENDFOR
\end{algorithmic}
\end{algorithm}

Inspired by the Representer Theorem \cite{scholkopf2001generalized}, which indicates that optimally trained classifiers can be depicted as linear combinations of their training samples, we concatenate the five inferred probabilities as a new input to a meta-linear MLP classifier for making the final prediction:

\begin{equation}
\label{eqa:5}
    \hat{y} = softmax(MLP(f_t \oplus f_m \oplus f_c \oplus f_{mt} \oplus f_{tm})
\end{equation}

Instead of optimizing modality-specific weights independently, linear classification through the proposed CMA simultaneously determines all weights to minimize the training loss. Consequently, we convert the standard n-shot classification to an $(n \times z)$-shot problem. The training details for CMA are presented in Algorithm 1.

\section{Experiment}
This section details experiments conducted to validate the effectiveness of the proposed approach. Initially, benchmark datasets are introduced, followed by the implementation details for experiments. The experimental results are analyzed in comparison to unimodal, multimodal, and few-shot FND methods. Finally, detailed analyses are provided to enhance the understanding of the proposed methods.

\subsection{Data Setup}
Three publicly available datasets are utilized for evaluation.

\textbf{PolitiFact} \cite{shu2020fakenewsnet} comprises a dataset of political news categorized as either fake or real by expert evaluators and is part of the benchmark FakeNewsNet project. Using the provided data crawling scripts, news with no images or invalid image URLs are removed, resulting in 198 multimodal news articles.

\textbf{GossipCop} \cite{shu2020fakenewsnet} features entertainment stories rated on a scale from 0 to 10, with stories scoring less than five classified as fake news by the author of FakeNewsNet. Using the same retrieval strategies as PolitiFact, 6,805 multimodal news articles are collected.

\textbf{Weibo} \cite{jin2017multimodal}, a dataset sourced from Chinese social media platforms, comprises a multimodal fake news collection featuring both text and images. Authentic news items were crawled from a reputable source (Xinhua News), and fake news was obtained from Weibopiyao, an official rumor refutation platform of Weibo, that aggregates content either through crowdsourcing or official rumor refutation efforts. The same pre-processing methods as in previous work \cite{wang2023cross} are followed, resulting in 7,853 Chinese news articles.

Notably, a news article might be accompanied by multiple images. To find the most relevant image, the cosine similarity between each image and its corresponding text is calculated, and the image-text pair with the highest similarity, as determined by the pre-trained CLIP, is retained. The resulting dataset statistics are presented in Table \ref{tab:1}.

\begin{table}[h!]
    \centering
    \caption{The statistics of the pre-processed multimodal fake news datasets. Avg tokens denote the average number of tokens per article.}
    \resizebox{0.41\textwidth}{!}{% Resize table to fit the text width
    \begin{tabular}{ c c c c } 
        \hline
        Statistics & PolitiFact & GossipCop & Weibo \\
        \hline
        Total news & 198 & 6,805 & 7,853\\ 
        Fake news & 96 &  1,877 & 4,211\\ 
        Real news &  102 & 4,928  & 3,642\\ 
        Avg tokens & 2,148 & 728 & 67\\
        \hline
    \end{tabular}
    }
    \label{tab:1}
\end{table}

\subsection{Implementation details}

The pre-trained OpenAI CLIP (ViT-B-32) \cite{radford2021learning} and Chinese CLIP (ViT-B-16) \cite{yang2022chinese} models are utilized to respectively extract text and image features for different languages. The hidden size for the cross-attention projection layer is 512, which is the same as the output dimension of CLIP encoders. The AdamW optimizer is employed with a learning rate of $1\mathrm{e}{-3}$ and a decay parameter of $1\mathrm{e}{-2}$. The model is trained for 20 epochs, with the optimal checkpoint being determined by peak validation performance. Early stopping is utilized with a patience of three epochs.

In the few-shot context, the model is trained using a restricted set of samples, selected from the dataset to form an n-shot scenario. Here, $n \in [2, 8, 16, 32]$ represents the number of samples for each class, while the remainder of the samples are reserved for testing purposes. Given that the data quality of the sampled training set might significantly impact the model’s performance, data sampling is repeated 10 times with random seeds, and the average score is reported after excluding the highest and lowest scores.

\subsection{Benchmarked Models}
The proposed CMA is benchmarked against 11 representative models. Specifically, we extensively compare the proposed method with unimodal approaches (1)-(3), multimodal approaches (4)-(6), and the few-shot approaches (7)-(11). 

(1) \textbf{dEFEND} \cite{10.1145/3292500.3330935} utilizes the hierarchical attention network for FND. In this study, we remove the user comments from the original model.

(2) \textbf{LDA-HAN} \cite{jiang2020comparing} integrates pre-calculated topic distributions from Latent Dirichlet Allocation into a hierarchical attention network for text classification.

(3) \textbf{FT-RoBERTa} is a standard, fine-tuned version of the pre-trained language model RoBERTa; we use Huggingface Trainer to conduct the fine-tuning experiment.

(4) \textbf{SpotFake} \cite{singhal2019spotfake} employs the pre-trained VGG and BERT for extracting image and text features, respectively, and then concatenating them for final classification. 

(5) \textbf{SAFE} \cite{zhou2020similarity} transforms images into textual descriptions and utilizes the correlation between text and visual information for FND.

(6) \textbf{CAFE} \cite{chen2022cross} employs an ambiguity-aware multimodal strategy to adaptively aggregate unimodal features and their correlations.

(7) \textbf{KPL} \cite{jiang2022fake} employs prompt learning in RoBERTa by enhancing it with external knowledge representations.

(8) \textbf{M-SAMPLE} \cite{JIANG2023119446} incorporates prompt learning with multimodal FND. It also applies a similarity-aware fusing to adaptively combine the intensity of multimodal representation for FND.

(9) \textbf{PET} \cite{schick2021exploiting} employs PLMs with task descriptions for supervised training, employing task-related cloze questions and verbalizers.

(10) \textbf{KPT} \cite{hu2022knowledgeable} enhances the label word space by incorporating class-related tokens that exhibit diverse granularities and perspectives.

(11) \textbf{P\&A} \cite{wu2023prompt} combines prompt-based learning with social alignment techniques and addresses label scarcity by using task-specific prompts in PLMs to elicit relevant knowledge.

\subsection{Results}
Table \ref{tab:performance} demonstrates the FND accuracy comparison between the proposed CMA and all the baselines at various few-shot settings over the three datasets. 

\begin{table*}[!htbp]
\centering
\caption{Performance comparison between the CMA and the baseline models in accuracy (\%). \textbf{Bold} indicates the best performance. \underline{Underline} is the second-best performance. \textbf{AVG} denotes the average accuracy per model across all n-shot settings and datasets. Notably, the experimental results of P\&A in Weibo are not accessible since it would require constructing the news proximity graph from the raw social context, which is not provided in the Weibo dataset.}
\label{tab:performance}
\resizebox{0.81\textwidth}{!}{% Resize table to fit the text width
\begin{tabular}{lccccccccccccc}
\toprule
\multirow{2}{*}{\textbf{Method}} & \multicolumn{4}{c}{\textbf{PolitiFact}} & \multicolumn{4}{c}{\textbf{GossipCop}} & \multicolumn{4}{c}{\textbf{Weibo}} &  \\
\cmidrule(lr){2-5} \cmidrule(lr){6-9} \cmidrule(lr){10-13}
& 2 & 8 & 16 & 32 & 2 & 8 & 16 & 32 & 2 & 8 & 16 & 32 & \textbf{AVG} \\
\midrule
% TextCNN &  &  &  &  &  &  &  &  &  &  &  &  \\
% BiLSTM &&&&&&&&&&&& \\
\textbf{dEFEND} & 21.3 & 39.7 & 37.5 & 54.1 & 25.6 & 26.0 & 44.1 & 47.8 & 31.9 & 33.0 & 40.1 & 44.5 & 37.1 \\
\textbf{LDA-HAN} & 39.4 & 47.3 & 52.2 & 54.9 & 21.2 & 30.4 & 39.5 & 41.3 & 40.3 & 41.8 & 44.4 & 50.9 & 42.0 \\
\textbf{FT-RoBERTa} & 52.0 & 63.1 & 70.0 & 72.5 & 41.3 & 60.4 & 62.6 & 65.9 & 39.7 & 58.1 & 64.3 & 66.3 & 59.7 \\
% ... Other rows here
\midrule
\textbf{SAFE} & 19.0 & 27.3 & 48.7 & 52.1 & 31.3 & 45.2 & 45.4 & 47.1 & 21.1 & 19.3 & 39.4 & 41.1 & 36.4 \\
\textbf{SpotFake} & 49.3 & 53.7 & 58.5 & 63.4 & 28.3 & 28.4 & 34.4 & 36.1 & 36.9 & 41.3 & 40.4 & 53.7 & 43.7 \\
\textbf{CAFE} & 38.6 & 46.4 & 48.9 & 51.0 & 42.3 & 48.1 & 55.9 & 59.3 & 44.4 & 40.6 & 47.5 & 51.3 & 47.9 \\
\midrule
\textbf{KPL} & 55.1 & 60.7 & 65.5 & 66.3 & 53.3 & 54.8 & 58.6 & 61.3 & 45.4 & 49.3 & 50.2 & 59.9 & 56.7\\
\textbf{M-SAMPLE} & 56.2 & 66.1 & 69.5 & 73.4 & 53.4 & 54.1 & 59.7 & 66.0 & 49.7 & 52.1 & 59.8 & 65.7 & 60.5 \\
\textbf{KPT} & 68.1 & 74.8 & 80.0 & 83.2 & 52.5 & 56.5 & 58.1 & 67.0 & 56.9 & \underline{69.4} & 69.9 & 71.2 & 67.3 \\
\textbf{PET} & \underline{73.2} & 68.4 & 68.3 & 70.1 & 65.7 & \underline{66.9} & 68.3 & \underline{71.1} & \underline{65.4} & 66.6 & \underline{70.3} & \underline{71.5} & 68.8 \\
\textbf{P\&A} & 71.9 & \textbf{80.7} & \underline{81.7} & \underline{83.5} & \underline{54.9} & 58.4 & \textbf{75.6} & 69.3& - & - & - & - & \underline{72.0} \\
\midrule
\textbf{CMA(Ours)} & \textbf{73.5} & \underline{75.8} & \textbf{82.5} & \textbf{87.3} & \textbf{71.9} & \textbf{69.0} & \underline{71.7} & \textbf{77.0} & \textbf{74.5} & \textbf{69.9} & \textbf{73.8} & \textbf{76.5} & \textbf{75.3} \\

\bottomrule
\end{tabular}
}
\end{table*}

\textbf{Comparing with unimodal baselines.} First, we assess the accuracy of both unimodal approaches and the proposed CMA to evaluate their performances. Overall, CMA outperforms the best unimodal approach, FT-RoBERTa, achieving a 15.6\% enhancement in average accuracy across all datasets, demonstrating its superiority in few-shot scenarios.

Surprisingly, FT-RoBERTa emerges as the most accurate model among both unimodal and multimodal approaches, suggesting that conventional fine-tuning methods can reach competitive levels of performance solely through the analysis of textual information from fake news. However, this method necessitates increased epoch time due to the adjustment of numerous parameters in the pre-trained language model (as shown in Table \ref{tab:params}), making it impractical for real-world few-shot FND applications.

LDA-HAN yields the second best in accuracy among unimodal models, with dEFEND coming in next. This could be attributed to two factors: firstly, the vanilla LDA model struggles to effectively generate topics from short texts, a characteristic of the datasets from GossipCop and Weibo (as detailed in Table \ref{tab:1}) used in LDA-HAN; secondly, the employment of GloVe embeddings for initializing LDA-HAN and dEFEND may not perform as effectively as the contextualized embeddings generated by the BERT family.

\textbf{Comparing with multimodal baselines.} We evaluate the performance of CMA in comparison with multimodal approaches. CMA outperforms the best multimodal baseline, CAFE, with a 27.4\% improvement in average accuracy across all datasets. The reason might be that the complex architecture of multimodal approaches inherently comes with a large number of trainable parameters, which might easily lead to overfitting in few-shot scenarios.

Excluding FT-RoBERTa, all multimodal baselines outperform unimodal models on average, showing that the inclusion of the image modality can significantly affect model accuracy. While these multimodal approaches excel in scenarios with abundant data, their effectiveness heavily relies on the availability of high-quality annotated training samples, which may not be readily accessible during the initial stages of FND. Moreover, all multimodal approaches utilize pre-trained unimodal models, such as VGG, ResNet, and BERT, to independently extract features from images and text. Yet, since these unimodal models are trained separately, merging their extracted features during the multimodal fusion process could potentially introduce noise\cite{JIANG2023119446}.

\textbf{Comparing with few-shot baselines.} The effectiveness of the proposed CMA is evaluated in comparison with the latest prompt-based few-shot models. CMA outperforms the best few-shot baseline, P\&A, with a 3.3\% improvement in average accuracy, showing that using unimodal features to assist multimodal probing without prompting the pre-trained language model could also benefit the FND task. 

While P\&A demonstrates performance on par with CMA, it requires the pre-calculation of a news proximity graph. However, such social context data may not always be accessible, particularly in datasets not sourced from Twitter, like Weibo. After analyzing PET and KPT, it's evident that these methods yield comparable outcomes, likely due to variations introduced by the manually crafted verbalizers used in prompting. This underscores the significance of hand-designed discrete templates in prompt-based learning. Concurrently, M-SAMPLE, a multimodal adaptation of KPL, demonstrates superior performance, suggesting that incorporating image modality can significantly enhance FND effectiveness.

\section{Analysis}

% This section conducts a comprehensive analysis of the CMA method in the few-shot setting. Initially, an ablation study is conducted to scrutinize the critical elements of CMA. Subsequently, the standard deviations of the model are presented. Finally, we visualize and contrast the embeddings derived from various baselines.

\subsection{Ablation study}
\label{sec:abl}
We investigate the impact of key components in CMA by assessing the framework's performance in a range of complete and partial configurations. In each experiment, CMA is selectively utilized by removing different components, followed by training the framework from scratch. The results are averaged over five random seeds in each shot, and indicate the performance decay of CMA in the absence of each component in most configurations, underscoring the significance of each key module within CMA, as shown in Table \ref{tab:ablation}.

\begin{table*}[!htbp]
\centering
\caption{Ablation experiments of the CMA. \textbf{-cross} denotes the cross-attention is removed from the CMA. \textbf{-meta} means the meta-linear MLP layer is removed. \textbf{-img} means the image features are removed and only text features are used. \textbf{-txt} denotes the text features are removed and only image features are used.}
\label{tab:ablation}
\resizebox{0.81\textwidth}{!}{% Resize table to fit the text width
\begin{tabular}{lccccccccccccc}
\toprule
\multirow{2}{*}{\textbf{Method}} & \multicolumn{4}{c}{\textbf{PolitiFact}} & \multicolumn{4}{c}{\textbf{GossipCop}} & \multicolumn{4}{c}{\textbf{Weibo}} &  \\
\cmidrule(lr){2-5} \cmidrule(lr){6-9} \cmidrule(lr){10-13}
& 2 & 8 & 16 & 32 & 2 & 8 & 16 & 32 & 2 & 8 & 16 & 32 & \textbf{AVG} \\
\midrule
% TextCNN &  &  &  &  &  &  &  &  &  &  &  &  \\
% BiLSTM &&&&&&&&&&&& \\
\textbf{CMA} & 73.5 & 75.8 & 82.5 & 87.3 & 71.9 & 69.0 & 71.7 & 77.0 & 74.5 & 69.9 & 73.8 & 76.5 & 75.3 \\
\textbf{-cross} & 67.6 & 76.7 & 81.2 & 84.0 & 71.8 & 71.8 & 71.6 & 71.1 & 58.4 & 65.2 & 68.4 & 75.2 & 71.6 \\
\textbf{-meta} & 72.2 & 74.1 & 74.7 & 78.4 & 49.0 & 53.2 & 56.8 & 56.1 & 50.0 & 50.9 & 57.4 & 61.7 & 61.2 \\

\textbf{-img} & 59.6 & 61.7 & 68.5 & 71.4 & 48.3 & 48.4 & 54.3 & 56.1 & 46.9 & 47.3 & 50.4 & 52.1 & 55.4 \\
\textbf{-txt} & 39.0 & 37.4 & 45.6 & 52.1 & 41.3 & 43.3 & 45.1 & 47.6 & 39.1 & 39.3 & 39.4 & 45.1 & 42.9 \\

% \textbf{CMA} & 0.735 & 0.758 & 0.825 & 0.873 & 0.719 & 0.690 & 0.717 & 0.770 & 0.745 & 0.699 & 0.738 & 0.765 & 0.753 \\
% \textbf{-cross} & 0.676 & 0.767 & 0.812 & 0.840 & 0.718 & 0.718 & 0.716 & 0.711 & 0.584 & 0.652 & 0.684 & 0.752 & 0.716 \\
% \textbf{-meta} & 0.722 & 0.741 & 0.747 & 0.784 & 0.490 & 0.532 & 0.568 & 0.561 & 0.500 & 0.509 & 0.574 & 0.617 & 0.612 \\

% \textbf{-img} & 0.596 & 0.617& 0.685 & 0.714 & 0.483 & 0.484& 0.543 & 0.561 & 0.469 & 0.473 & 0.504 & 0.521 & 0.554 \\
% \textbf{-txt} & 0.390 & 0.374 & 0.456 & 0.521 & 0.413 & 0.433 & 0.451 & 0.476 & 0.391 & 0.393 & 0.394 & 0.451 & 0.429 \\

\bottomrule
\end{tabular}
}
\end{table*}

Specifically, removing the cross-attention from the CMA (i.e., \textbf{-cross}) results in a slight decrease in accuracy, showing that the cross-attended features from text and image capture semantic correlations and contribute to improved performance. Further removal of the meta-linear layer from the CMA (i.e., \textbf{-meta}) transforms the model into a standard n-shot classification, where it simply classifies concatenated multimodal features. This leads to a significant decrease in accuracy, emphasizing the importance of jointly updating all modality-specific weights in a meta-linear classifier for cross-modal adaptation and accuracy improvement. The meta-linear layer integrates modality-specific features, resembling an ensemble that transforms n-shot classification into a more robust $(n \times z)$-shot problem, enhancing cross-modal adaptation in few-shot classification.

Additionally, experiments are performed by excluding either the image features (\textbf{-img}) or the text features (\textbf{-txt}), relying solely on the remaining modality for classification. Such setups led to additional reductions in accuracy, underscoring the comparative importance of text over image features in FND. This highlights the complexities in multimodal FND tasks, where the spatial discrepancies between visual and textual semantics tend to be more subtle than in broader multimodal datasets.
 
\subsection{Stablility test}

\begin{figure}[h!]
     \centering
     \begin{subfigure}[b]{0.45\textwidth}
         \centering
         \includegraphics[width=\textwidth]{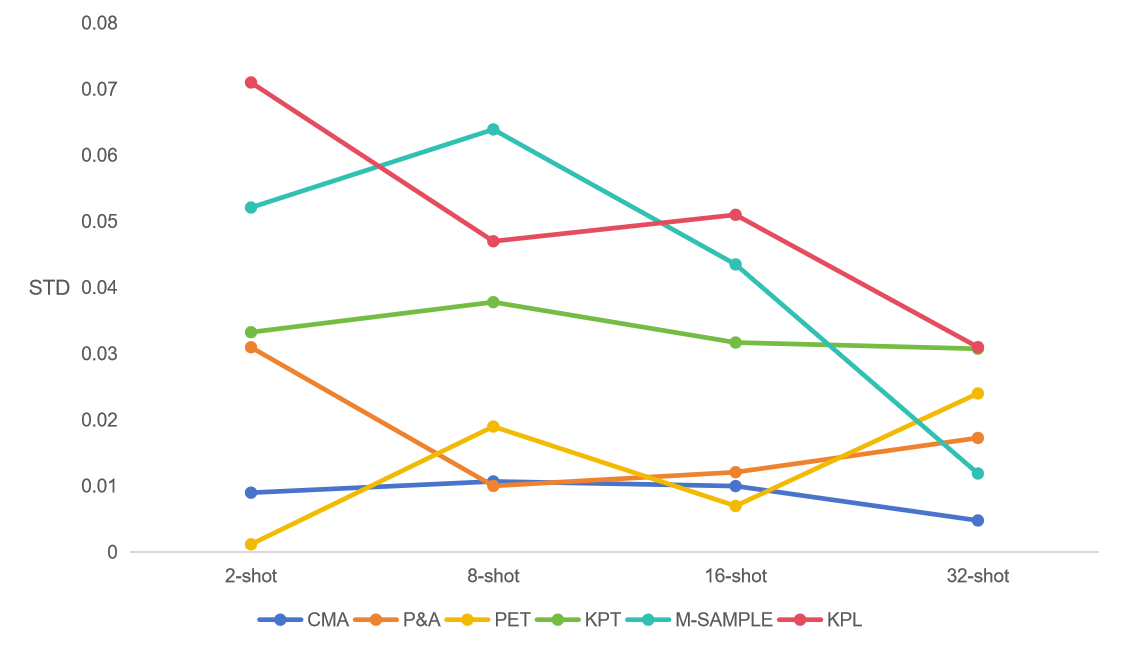}
         \caption{Standard deviation comparisons in the PolitiFact.}
         \label{FIG:std1}
     \end{subfigure}
     \begin{subfigure}[b]{0.45\textwidth}
         \centering
         \includegraphics[width=\textwidth]{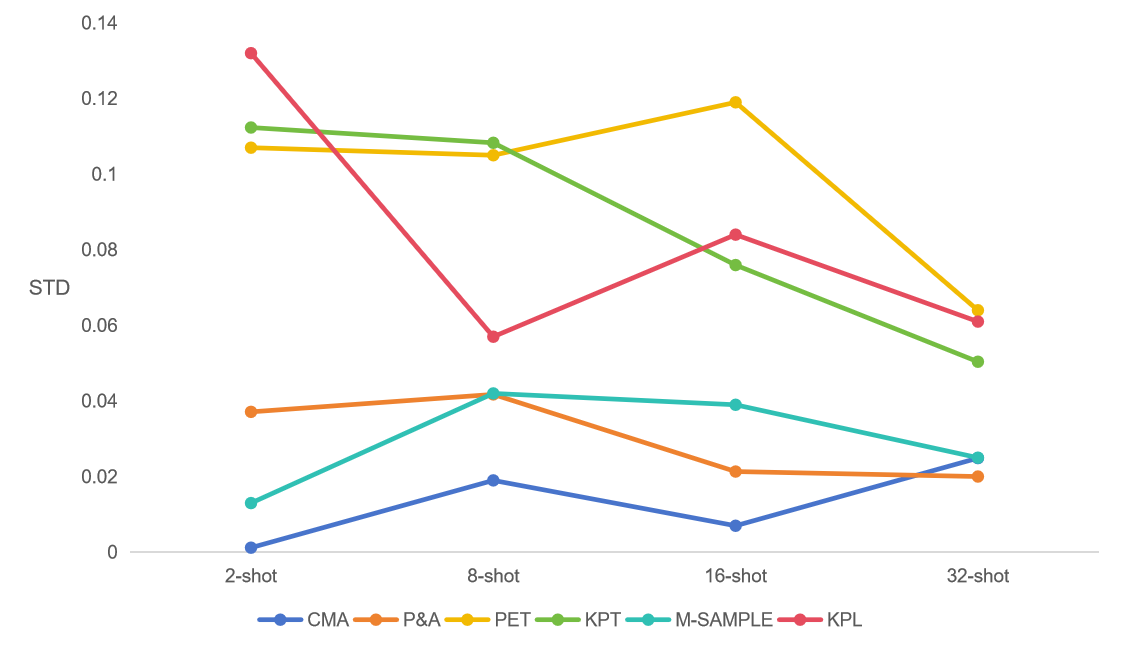}
         \caption{Standard deviation comparisons in the GossipCop.}
         \label{FIG:std2}
     \end{subfigure}
        \caption{The standard deviations of accuracies for both PolitiFact and GossipCop datasets among the few-shot baselines and the proposed CMA.}
        \label{FIG:std}
\end{figure}

Given the selection of few-shot examples can significantly affect the model performance, we assess the stability of the CMA and other prompt-based baselines by measuring the standard deviation of accuracies in the few-shot settings, as shown in Figure \ref{FIG:std}.

Overall, the standard deviation for all models decreases in tandem with an increase in the number of n-shot settings, underscoring the importance of augmenting training examples in few-shot scenarios. This augmentation can be further observed that the standard deviation of the CMA tends to be the most stable among the few-shot approaches, indicating that the ensemble of unimodal features in the meta-linear layer can enhance the robustness of multimodal fusion in classification. Additionally, the GossipCop dataset exhibits greater instability compared to the PolitiFact dataset. This instability may be attributed to the semantic complexity in GossipCop, which is responsible for the lower accuracy across all models.

\subsection{Model efficiency}

Given the CMA achieves the best performance with a surprisingly simple augmentation, we further explore its efficiency in comparison to other baseline models. Table \ref{tab:params} showcases a comparison of the accuracies and epoch times between baselines and the CMA. The average accuracy of each model is determined in a 16-shot setting as shown in Table \ref{tab:performance}, along with the recording of average epoch times for each model. All experiments are tested with batch size 32 on a single RTX 4090 GPU in the GossipCop dataset for a fair comparison.

% Model & avg\_acc &num\_params & epoch\_time  \\
% \midrule
% dEFEND & 0.359 &  0.17M & \textcolor{green}{3min} \\
% LDA-HAN & 0.331 & 0.17M & 8min   \\
% FT-RoBERTa & 0.576 & 125M & \textcolor{red}{10min} \\
% \hline
% SAFE & 0.423 & 0.12M & \textcolor{red}{21min} \\
% Spotfake & 0.318 & 13M & \textcolor{green}{2min} \\
% CAFE & 0.514 & 0.95M & 3min \\
% \hline
% KPL & 0.570 & 0.68M & 5min \\
% M-SAMPLE & 0.583 & 0.64M & 7min \\
% KPT & 0.585 & 355M & \textcolor{green}{3min} \\
% PET & 0.680 & 356M & \textcolor{red}{14min} \\
% P\&A & 0.646 & 109M & \textcolor{green}{2min} \\
% CMA & 0.724 & 0.79M & \textcolor{green}{$<$1min} \\

% Considering the semantic complexity and the relatively large size of the dataset in GossipCop, we calculate the average accuracy, trainable parameters, and the time elapsed in each epoch for the n-shot settings between the proposed CMA and the baselines.
Among unimodal models, dEFEND and LDA-HAN exhibit comparable accuracy and epoch times, attributed to their analogous hierarchical architectural design. While FT-RoBERTa exceeds the performance of various unimodal (e.g., 18\% higher than dEFEND) and multimodal methods (e.g., 6.9\% higher than CAFE), it requires modifying a significant number of trainable parameters, thus extending epoch durations (on average, four minutes per epoch) relative to other unimodal baselines.

In the multimodal models, SAFE yields the lengthiest epoch durations owing to its prerequisite for independently pre-generating image descriptions. Although Spotfake achieves the fastest epoch duration due to its simple concatenation of the image and text features from the BERT and VGG respectively, it achieves the worst performance compared with other models. CAFE achieves the best multimodal FND outcomes by integrating a degree of ambiguity in the similarity across text and image features, albeit at the cost of marginally increased model complexity and consequently, slightly extended epoch durations.

\begin{table}[!h]
\caption{Comparisons of model efficiency. Both Accuracy (\%) and Time represent averages derived from five random seeds. Times displayed in \textcolor{green}{green} signify an average duration of less than 3 minutes, whereas those in \textcolor{red}{red} indicate an average exceeding 3 minutes. Gain denotes notable improvements in accuracy relative to the dEFEND model.}\label{tab:7}
\centering
\resizebox{0.36\textwidth}{!}{% Resize table to fit the text width
\begin{tabular}{lccc}
\toprule
Model & Accuracy & Time & Gain  \\
\midrule
% dEFEND & 0.409 &  \textcolor{green}{2min} & 0 \\
% LDA-HAN & 0.387 &  \textcolor{green}{2.5min} & -0.022  \\
% FT-RoBERTa & 0.589 &  \textcolor{red}{4min} & +0.180 \\
% \hline
% SAFE & 0.411 &  \textcolor{red}{7min} & +0.002 \\
% Spotfake & 0.339 &  \textcolor{green}{2min} & -0.070 \\
% CAFE & 0.520 &  3min & +0.111 \\
% \hline
% KPL & 0.575 &  3min & +0.166 \\
% M-SAMPLE & 0.581 &  \textcolor{red}{5min} & +0.172 \\
% KPT & 0.543 &  3min & +0.134 \\
% PET & 0.699 &  \textcolor{red}{6min} & +0.290 \\
% P\&A & 0.715 &  \textcolor{green}{2min} & +0.306  \\
% CMA & 0.741 &  \textcolor{green}{$<$1min} & +0.332 \\

\textbf{dEFEND} & 40.9 &  \textcolor{green}{2min} & 0 \\
\textbf{LDA-HAN} & 38.7 &  \textcolor{green}{2min} & -2.2  \\
\textbf{FT-RoBERTa} & 58.9 &  \textcolor{red}{4min} & +18.0 \\
\hline
\textbf{SAFE} & 41.1 &  \textcolor{red}{7min} & +0.2 \\
\textbf{Spotfake} & 33.9 &  \textcolor{green}{2min} & -7.0 \\
\textbf{CAFE} & 52.0 &  3min & +11.1 \\
\hline
\textbf{KPL} & 57.5 &  3min & +16.6 \\
\textbf{M-SAMPLE} & 58.1 &  \textcolor{red}{5min} & +17.2 \\
\textbf{KPT} & 54.3 &  3min & +13.4 \\
\textbf{PET} & 69.9 &  \textcolor{red}{6min} & +29.0 \\
\textbf{P\&A} & 71.5 &  \textcolor{green}{2min} & +30.6  \\
\textbf{CMA} & 74.1 &  \textcolor{green}{$<$1min} & +33.2 \\
\bottomrule
\label{tab:params}

\end{tabular}
}
\end{table}

All few-shot baselines demonstrate significant improvements over both unimodal and multimodal counterparts, indicating the suboptimality of traditional methods in contexts with limited annotated data. Specifically,  the integration of external knowledge into the prompt-tuning phase by both KPL and KPT results in comparable epoch durations. However, KPL's design of an FND-specific prompt may underlie its superior performance over KPT. PET records the lengthiest epoch duration among the few-shot baselines, potentially due to the repeated fine-tuning of the PLM for reconfiguring input examples with the task description. P\%A not only achieves the second-best performance but also the second-shortest epoch durations, benefiting from the integration of user engagements. However, it incorporates an external alignment module to correlate user engagement with the PLM's predictions, consequently increasing epoch times relative to CMA. Finally, CMA is more efficient and precise as it avoids the need for extensive parameter fine-tuning and does not depend on intensive image augmentation processes. Additionally, the inclusion of linear probing layers atop the image and text features presents a more streamlined approach than extensive fine-tuning and precise-crafted complex model designs.

% Meanwhile, prompt-based approaches like P\&A, PET, and KPT, which unfreeze the parameters of language models, result in a larger number of trainable parameters compared to frozen ones (i.e., M-SAMPLE and KPL). However, freezing the language model also leads to a decrease in accuracy.

\subsection{Domain shift analysis}

Real-world fake news demonstrates significant distribution discrepancies, which is also referred to as domain shift \cite{zhu2022generalizing, zhu2022memory}. Consequently, automatic FND methods are required to rapidly adapt to emerging topics by using limited resources.

\begin{table}[!ht]
\centering
\caption{Domain shift performance comparison. Poli$\rightarrow$Goss refers to utilze few-shot samples from the Politifact as training and the Gossipcop for testing. Goss$\rightarrow$Poli denotes the Gossipcop is utilized as training set and the Politifact is the test set. Bold and \underline{Underline} denote the best and the second best accuracy (\%) in that n-shot setting. AVG is the mean accuracy across all n-shot settings.}
\label{tab:domain}
\resizebox{0.48\textwidth}{!}{% Resize table to fit the text width
\begin{tabular}{lccccccccc}
\toprule
\multirow{2}{*}{\textbf{Method}} & \multicolumn{4}{c}{\textbf{Poli$\rightarrow$Goss}} & \multicolumn{4}{c}{\textbf{Goss$\rightarrow$Poli}}  &  \\
\cmidrule(lr){2-5} \cmidrule(lr){6-9}
& 2 & 8 & 16 & 32 & 2 & 8 & 16 & 32  & \textbf{AVG} \\
\midrule

\textbf{KPT} & 40.1 & 31.7 & 31.4 & 31.1 & \textbf{56.3} & \textbf{55.3} & \underline{54.1} & \underline{55.8} & 44.5 \\
\textbf{PET} & \underline{51.0} & 51.3 & 51.5 & 51.6 & \underline{53.1} & \underline{54.1} & \textbf{54.5} & 54.1 & \underline{52.6} \\
\textbf{P\&A} & \textbf{53.2} & \underline{53.4} & \underline{53.2} & \underline{54.5} & 50.1 & 50.4 & 50.3 & 50.5 & 51.9 \\
\textbf{CMA} & 48.7 & \textbf{53.5} & \textbf{56.1} & \textbf{58.6} & 51.4 & \textbf{55.3} & 53.0 & \textbf{55.9} & \textbf{54.1} \\

% \textbf{KPT} & 0.401 & 0.317 & 0.314 & 0.311 & \textbf{0.563} & \textbf{0.553} & \underline{0.541} & \underline{0.558} & 0.445 \\
% \textbf{PET} & \underline{0.510} & 0.513 & 0.515 & 0.516 & \underline{0.531} & \underline{0.541} & \textbf{0.545} & 0.541 & \underline{0.526} \\
% \textbf{P\&A} & \textbf{0.532} & \underline{0.534} & \underline{0.532} & \underline{0.545} & 0.501 & 0.504 & 0.503 & 0.505 & 0.519 \\
% \textbf{CMA} & 0.487 & \textbf{0.535} & \textbf{0.561} & \textbf{0.586} & 0.514 & \textbf{0.553} & 0.530 & \textbf{0.559} & \textbf{0.541}  \\

\bottomrule
\end{tabular}
}
\end{table}

To address this, we investigate the cross-domain capability of the proposed CMA against three strong few-shot FND baselines (i.e., P\&A, PET and KPT). Considering Politifact's focus on political news using formal language and Gossipcop's emphasis on entertainment and celebrity narratives in a more casual tone, we first utilize Politifact for training and Gossipcop for testing, later inverting this arrangement.

The outcomes following domain shift are presented in Table \ref{tab:domain}. Notably,while the CMA model records the highest average accuracy among the few-shot baselines, the performance of each model markedly differs from that observed in the comparison experiments (as shown in Table \ref{tab:performance}). For example, KPT exhibits the strongest performance in both 2- and 8-shot scenarios in Goss$\rightarrow$Poli. PET and P\%A also achieve the highest performance in Goss$\rightarrow$Poli and Poli$\rightarrow$Goss respectively, highlighting the disparity between present few-shot FND methodologies and their adaptability to domain adaptation.

\subsection{Feature visualization}

\begin{figure}[h!]
     \centering
     \begin{subfigure}[b]{0.47\textwidth}
         \centering
         \includegraphics[width=\textwidth]{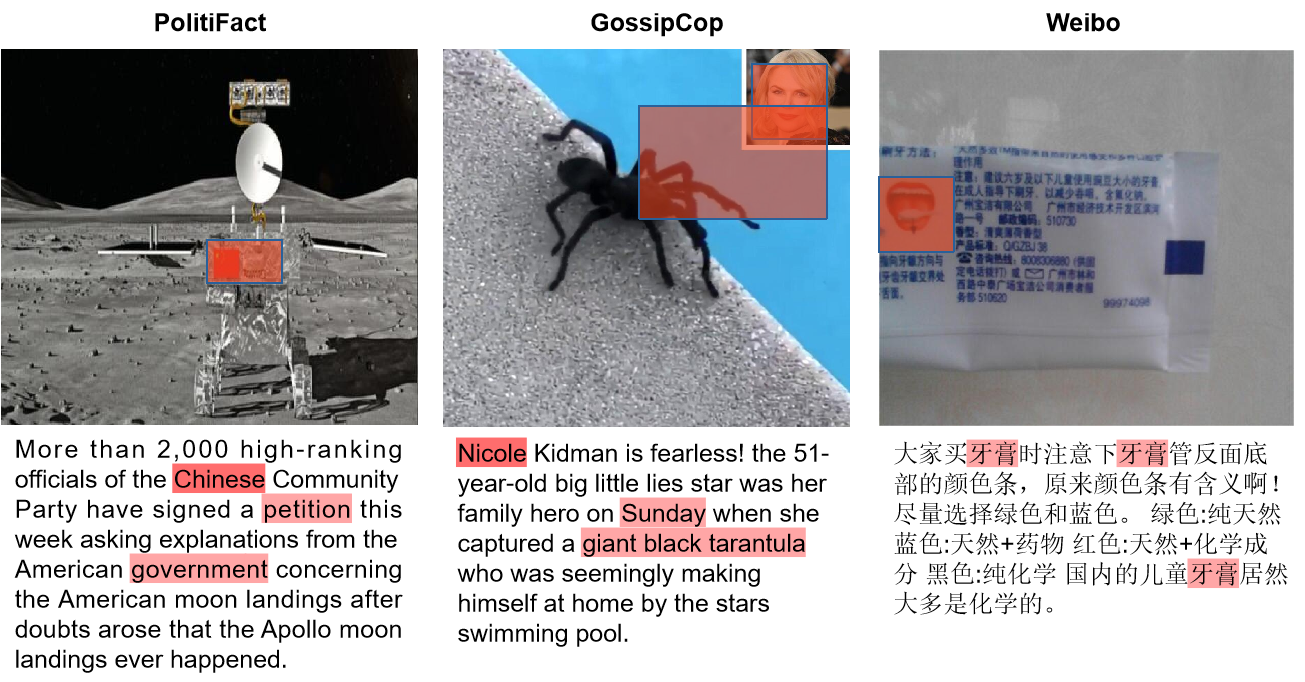}
         \caption{Feature visualization from the M-SAMPLE.}
         \label{FIG:vis1}
     \end{subfigure}
     \begin{subfigure}[b]{0.47\textwidth}
         \centering
         \includegraphics[width=\textwidth]{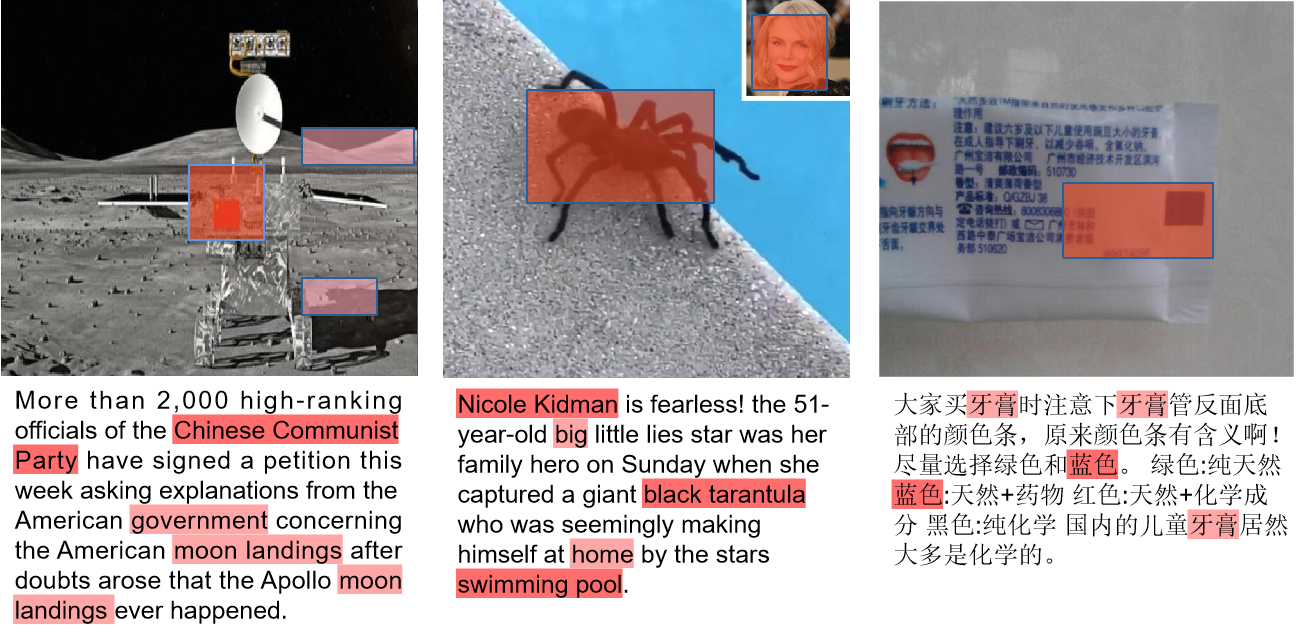}
         \caption{Feature visualization from the proposed CMA.}
         \label{FIG:vis2}
     \end{subfigure}
        \caption{Feature visualization comparisons between M-SAMPLE and CMA. English translation of the Weibo example: ``\textit{When you buy toothpaste, pay attention to the color bar on the bottom of the toothpaste tube, the color bar has meaning! Try to choose greens and blues. Green: natural, blue: natural + medicine, Red: natural + chemical composition, Black: pure chemical. Surprisingly, most children's toothpaste brands on the domestic market contain chemical ingredients.}''
}
        \label{FIG:vis_all}
\end{figure}

At last, we present a visual comparison of the features extracted by M-SAMPLE and CMA, both of which are multimodal few-shot approaches. This involves the visualization of multimodal features alongside an assessment of their semantic correlations. For each dataset, a specific sample is chosen, with the corresponding multimodal features depicted in Figure \ref{FIG:vis_all}.

Observations indicate that: 1) CMA can capture more consistent features from the image-text pair of fake news than those of M-SAMPLE. For example, although both M-SAMPLE and CMA successfully correlate the flag in the image with the word ``Chinese'' in the text, CMA can also identify the semantic meaning of ``moon landing'' between the text and image in the PolitiFact example; 2) The proposed CMA is more accurate in capturing important features from the image than M-SAMPLE. For example, although both models can identify the person ``Nicole Kidman'' and ``black tarantula'' in both the text and the image in the GossipCop example, the image region of the tarantula slightly overlaps with that of Nicole Kidman provided by M-SAMPLE. This is even more obvious in the Weibo example, as CMA successfully captures the ``blue'' color bar in the toothpaste, but M-SAMPLE fails to do so.

\section{Conclusion}

This paper introduced Cross-Modal Augmentation (CMA) for enhancing few-shot multimodal fake news detection by utilizing unimodal features to augment multimodal fusion. The proposed CMA leverages a pre-trained multimodal model for unimodal feature extraction and transforms n-shot classification into a robust (n $\times$ z)-shot problem using class labels as additional one-shot training samples. The CMA, employing a simple linear classifier, achieves SOTA performance on three datasets in few-shot settings, and demonstrates greater efficiency than current approaches.

% The effectiveness of CMA in capturing nuanced semantic correlations between text and image in fake news, while remaining computationally efficient, is demonstrated through extensive experiments and analyses. The proposed CMA highlights the value of integrating multimodal representations, similar to human concept learning, in the domain of automated fake news detection.

\section{Limitation}

We acknowledge limitations in this study including: 1) The evaluation of CMA's few-shot proficiency solely utilizes CLIP, future investigations will delve into how different multimodal models influence the proposed CMA; 2) Given the lack of multimodal information in certain datasets, this research adopted cosine similarity for image selection from multiple options, potentially leading to varied performance outcomes based on the text-image pairing technique employed; 3) CMA exhibits suboptimal domain shift performance, enhancing the architecture through the integration of knowledge distillation or domain adaptation techniques remains a prospect for future research.

\section*{Acknowledgements}
This work is funded by the Natural Science Foundation of Shandong Province under grant ZR2023QF151 and the Natural Science Foundation of China under grant 12303103.

%and partially supported by a European Union Horizon 2020 Program under the scheme “INFRAIA-01-2018-2019 – Integrating Activities for Advanced Communities”, Grant Agreement n.871042 (“SoBigData++: European Integrated Infrastructure for Social Mining and Big Data Analytics” (http://www.sobigdata.eu)).

\section*{CRediT authorship contribution statement}

\noindent\textbf{Ye Jiang}: Conceptualization, Methodology, Writing–original draft, Writing–review \& editing. \textbf{Taihang Wang}: Methodology, Writing–review \& editing. \textbf{Xiaoman Xu}: Data curation, Writing – review \& editing. \textbf{Yimin Wang}: Funding acquisition, Methodology, Writing–review \& editing. \textbf{Xingyi Song}: Supervision, Writing – review \& editing. \textbf{Diana Maynard}: Investigation, Supervision, Writing – review \& editing.



\bibliographystyle{elsarticle-num} 
\bibliography{elsarticle-bib}

% \appendix





% \end{thebibliography}
\end{document}